%% file: arxiv_paper.tex
\documentclass[runningheads]{llncs}
\usepackage{graphicx}
\usepackage{comment}
\usepackage{bbold}
\usepackage{amsmath,amssymb} 
\usepackage{color, soul}
\usepackage{subcaption}

\usepackage{multirow}
\usepackage{titletoc}
\usepackage{graphicx}
\usepackage{multirow}
\usepackage[table,xcdraw]{xcolor}
\usepackage{booktabs}
\usepackage{arydshln}

\usepackage[thinc]{esdiff}
\usepackage{tablefootnote}
\usepackage{tabularx}
\usepackage{threeparttable}
\usepackage{footnote}
\usepackage{caption,pdfpages}
\usepackage{hyperref}
\usepackage[ruled]{algorithm2e}
\usepackage{minitoc}
\usepackage{blindtext}

\usepackage{listings}
\DeclareFixedFont{\ttb}{T1}{txtt}{bx}{n}{6} 
\DeclareFixedFont{\ttm}{T1}{txtt}{m}{n}{6}  

\definecolor{deepblue}{rgb}{0,0,0.5}
\definecolor{deepred}{rgb}{0.6,0,0}
\definecolor{deepgreen}{rgb}{0,0.5,0}

\definecolor{codegreen}{rgb}{0,0.6,0}
\definecolor{codegray}{rgb}{0.5,0.5,0.5}
\definecolor{codepurple}{rgb}{0.58,0,0.82}
\definecolor{backcolour}{rgb}{0.95,0.95,0.92}

\definecolor{codeblue}{rgb}{0.25,0.5,0.5}
\newcommand\beforecaptions{\vspace{-3mm}}
\newcommand\aftercaptions{\vspace{-5mm}}

\providecommand{\ie}[0]{\emph{i.e.}}
\providecommand{\eg}[0]{\emph{e.g.}}

\newcommand{\beginsupplement}{%
        \setcounter{table}{0}
        \renewcommand{\thetable}{S\arabic{table}}%
        \setcounter{figure}{0}
        \renewcommand{\thefigure}{S\arabic{figure}}%
     }
     
        \makeatletter
        \renewcommand*\l@author[2]{}
        \renewcommand*\l@title[2]{}
        \makeatletter

\begin{document}
\pagestyle{headings}
\mainmatter
\def\ECCVSubNumber{929}  

\title{Smooth-AP: Smoothing the Path \\ Towards Large-Scale Image Retrieval} 

\titlerunning{Smooth-AP: Smoothing the Path Towards Large-Scale Image Retrieval}
%
\authorrunning{A. Brown \textit{et al.}}
\author{Andrew Brown, Weidi Xie, Vicky Kalogeiton, Andrew Zisserman}
\institute{Visual Geometry Group, University of Oxford\\
\email{\{abrown,weidi,vicky,az\}@robots.ox.ac.uk} \\
\url{https://www.robots.ox.ac.uk/~vgg/research/smooth-ap/}}

\maketitle

\begin{abstract}
Optimising a ranking-based metric, such as Average Precision (AP), 
is notoriously challenging due to the fact that it is non-differentiable, 
and hence cannot be optimised directly using gradient-descent methods. 
To this end, we introduce an objective that optimises instead a~{\em smoothed approximation} of AP, coined {\em Smooth-AP}.  
Smooth-AP is a plug-and-play objective function that allows for end-to-end training of deep networks with a simple and elegant implementation. 
We also present an analysis for why directly optimising the ranking based metric of AP offers benefits over other deep metric learning losses.

We apply Smooth-AP to standard retrieval benchmarks : 
Stanford Online products and VehicleID,
and also evaluate on larger-scale datasets: 
INaturalist for fine-grained category retrieval, and VGGFace2 and IJB-C for face retrieval. 
In all cases, 
we improve the performance over the state-of-the-art, 
especially for larger-scale datasets, 
thus demonstrating the effectiveness and scalability of Smooth-AP to real-world scenarios. 
\end{abstract}

\section{Introduction}
\input{arxiv_sec/intro.tex}

\label{sec:intro}

\section{Related Work}
\input{arxiv_sec/related_work.tex}

\label{sec:related}

\section{Background}
\input{arxiv_sec/background.tex}
\label{sec:background}

\section{Approximating Average Precision~(AP)}
\input{arxiv_sec/method.tex}

\label{sec:method}

\section{Experimental Setup}
\input{arxiv_sec/experiment.tex}

\label{sec:experimental_setup}


\section{Results}
\input{arxiv_sec/results.tex}

\label{sec:experimental_results}

\section{Conclusions}
\input{arxiv_sec/conclusion.tex}
\label{sec:conclusion}

\clearpage
%
%

\bibliographystyle{splncs04}
\bibliography{shortstrings,vgg_local,vgg_other,sup_bib}

\title{Smooth-AP: Smoothing the Path \\ Towards Large-Scale Image Retrieval \\ \textit{Supplementary Material}} 

\titlerunning{Smooth-AP: Supplementary Material}

\authorrunning{A. Brown \textit{et al.}}
\author{Andrew Brown\orcidID{0000-0002-9556-2633} \and 
Weidi Xie\orcidID{0000-0003-3804-2639} \and \\
Vicky Kalogeiton\orcidID{0000-0002-7368-6993}\and \\
Andrew Zisserman\orcidID{0000-0002-8945-8573}}
\institute{Visual Geometry Group, University of Oxford\\
\email{\{abrown,weidi,vicky,az\}@robots.ox.ac.uk} \\
\url{https://www.robots.ox.ac.uk/~vgg/research/smooth-ap/}}


\beginsupplement

\maketitle

\begin{center}
  \textbf{\large{Table Of Contents}}
\end{center}
\startcontents[sections]
\printcontents[sections]{l}{1}{\setcounter{tocdepth}{1}}


\section{Further qualitative results}
\input{arxiv_sec/E.tex}

\label{sec:qual_results}
\newpage

\section{Source code}
Here, we provide pseudocode for the Smooth-AP loss written in PyTorch style. 
The simplicity of the method is demonstrated by the short implementation. 

\input{arxiv_sec/pseudo_code.tex}


\section{Details on the effects of increasing the mini-batch size on Smooth-AP}
\input{arxiv_sec/C.tex}
\label{sec:method}

\section{Choice of hyper-parameters for the compared-to methods for the INaturalist experiments}
\input{arxiv_sec/D.tex}
\label{sec:method}

\section{Complexity of the proposed loss}
\input{arxiv_sec/A.tex}

\label{sec:related}

%
%
\end{document}

%% file: arxiv_sec/intro.tex
Our objective in this paper is to improve the performance of 
`query by example', where the task is:
given a query image, rank all the instances in a retrieval set according to their relevance to the query. For instance, imagine that you have a photo of a friend or family member, and want
to search for all of the images of that person within your large
smart-phone image collection; or on a photo licensing site, you want
to find all photos of a particular building or object, starting from a
single photo. These use cases, where high recall is premium, differ
from the `Google Lens' application of identifying an object from an
image, where only one `hit' (match) is sufficient. 

The benchmark metric for retrieval quality is Average Precision (AP) (or its generalized variant, Normalized Discounted Cumulative Gain, which includes non-binary relevance judgements).  
With the resurgence of deep neural networks, 
end-to-end training has become the \emph{de facto} choice for solving specific vision tasks with well-defined metrics. 
However, the core problem with AP and similar metrics is that they include a discrete ranking function that
is neither differentiable nor decomposable. 
Consequently, their direct optimization, \eg\ with gradient-descent methods, 
is notoriously difficult. 

\begin{figure}[t!]
\begin{center}
   \includegraphics[width=\linewidth]{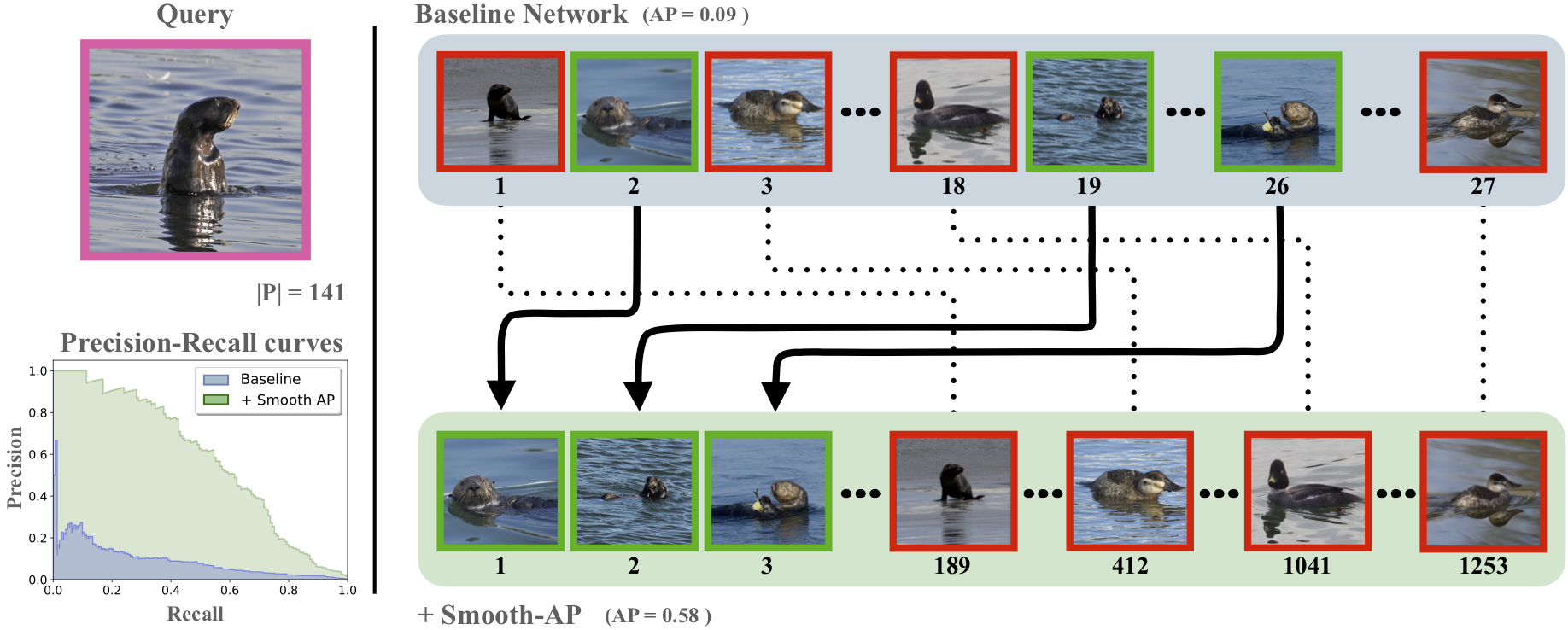}
\end{center}

\caption{\small{
\textbf{Ranked retrieval sets} before (top) and after (bottom) applying Smooth-AP on a baseline network (\ie\ ImageNet pre-trained weights) for a given query (pink image). 
The precision-recall curve is shown on the left. Smooth-AP results in large boost in AP, as it moves positive instances (green) high up the ranks and negative ones (red) low down. $|\mathcal{P}|$ is the number of positive instances in the retrieval set for this query. Images are from the INaturalist dataset.
}}

\label{fig:teaser}

\end{figure}

In this paper, we introduce a novel differentiable AP approximation,  
{\em Smooth-AP}, 
that allows end-to-end training of deep networks for ranking-based tasks. 
Smooth-AP is a simple, elegant, and scalable method that takes the form of a plug-and-play objective function 
by relaxing the Indicator function in the non-differentiable AP with a sigmoid function.
To demonstrate its effectiveness, 
we perform experiments on two commonly used image retrieval benchmarks, Stanford Online Products and VehicleID, 
where Smooth-AP outperforms all recent AP approximation approaches~\cite{Cakir19,Rolnek20optimizing} 
as well as recent deep metric learning methods.
We also experiment on three further large-scale retrieval datasets~(VGGFace2, IJB-C, INaturalist), 
which are orders of magnitude larger than the existing retrieval benchmarks. 
To our knowledge, 
this is the first work that demonstrates the possibility of training networks for AP on datasets 
with millions of images for the task of image retrieval.
We show large performance gains over all recently proposed AP approximating approaches 
and, somewhat surprisingly, 
also outperform strong verification systems~\cite{Deng19,Liu17} by a significant margin, 
reflecting the fact that metric learning approaches are indeed inefficient for training large-scale retrieval systems that are measured by global ranking metrics.

%% file: arxiv_sec/related_work.tex
As an essential component of information retrieval~\cite{Manning08}, 
algorithms that optimize rank-based metrics have been the focus of extensive research over the years.
In general, the previous approaches can be split into two lines of research, 
namely metric learning, and direct approximation of Average Precision.

\noindent {\bf Image Retrieval.} 
This is one of the most researched topics in the vision community.
Several themes have been explored in the literature,
for example, 
one theme is on the speed of retrieval
and explores methods of approximate nearest neighbors~\cite{Chum11,Jegou08,Jegou11,Jegou11b,Philbin07,Sivic03}.
Another theme is on how to obtain a compact image descriptor for retrieval in order to reduce the memory footprint.
Descriptors were typically constructed through an aggregation of local features, 
such as Fisher vectors~\cite{Perronnin10} and VLAD~\cite{Arandjelovic2013,Jegou10}.
More recently, neural networks have made impressive progress on learning representations for image retrieval~\cite{Arandjelovic16,Babenko14,Gordo16,Radenovic16,Wang14_deepranking},
but common to all is the choice of the loss function used for training; 
in particular, it should ideally be a loss that will encourage ‘good’ ranking.

\noindent {\bf Metric Learning.}
To avoid the difficulties from directly optimising rank-based metrics, such as Average Precision, 
there is a great body of work that focuses on metric learning \cite{Arandjelovic16,Arandjelovic2013,burges2007learning,Cao07,Chen19,Chopra05,Jegou11b,Law17,Movshovitz17,Hyun16,rao2018cvpr,Ustinova16,Wang19ranked,Weinberger06}.  For instance, the contrastive~\cite{Chopra05} and triplet~\cite{Weinberger06} losses, which consider pairs or triplets of elements,
and force all positive instances to be close in the high-dimensional embedding space, 
while separating negatives by a fixed distance (margin).
However, due to the limited rank-positional awareness that a pair/triplet provides, 
a model is likely to waste capacity on improving the order of positive instances 
at low (poor) ranks at the expense of those at high ranks, 
as was pointed out by Burges \textit{et al.} \cite{burges2007learning}.
Of more relevance,
the list-wise approaches~\cite{burges2007learning,Cao07,Movshovitz17,Hyun16,Wang19ranked}
look at many examples from the retrieval set, and have been proven to improve training efficiency and performance.
Despite being successful, one drawback of metric learning approaches is that
they are mostly driven by minimizing distances, and 
therefore remain ignorant of the importance of shifting ranking orders -- the latter is essential when evaluating with a rank-based metric.

\noindent {\bf Optimizing Average Precision~(AP).} 
The trend of directly optimising the non-differentiable AP has been recently revived in the retrieval community.
Sophisticated methods~\cite{Cakir19,Chen19,engilberge2019sodeep,he2018hashing,He18_ap,henderson2016end,mcfee2010metric,prillo2020softsort,Revaud19,Rolnek20optimizing,Song16,Taylor08,Ustinova16,cross_batch_MAP,Yue07}
have been developed to overcome the challenge of non-decomposability and non-differentiability in optimizing AP. 
Methods include:  
creating a distribution over rankings by treating each relevance score as a Gaussian random variable~\cite{Taylor08}, 
loss-augmented inference~\cite{mcfee2010metric}, 
direct loss minimization~\cite{henderson2016end,Song16}, 
optimizing a smooth and differentiable upper bound of AP~\cite{mcfee2010metric,Mohapatra18,Yue07},
training a LSTM to approximate the discrete ranking step~\cite{engilberge2019sodeep}, 
differentiable histogram binning~\cite{Cakir19,he2018hashing,He18_ap,Revaud19,Ustinova16},
error driven update schemes~\cite{Chen19b}, 
and the very recent blackbox optimization~\cite{Rolnek20optimizing}.
Significant progress on optimizing AP was made by the information retrieval community~\cite{burges2007learning,Chapelle07,John08,li2014maximal,Qin10,Taylor08}, but the methods have largely been ignored by the vision community, 
possibly because they have never been demonstrated on large-scale image retrieval or due to the complexity of the proposed smooth objectives. 
One of the motivations of this work is to show that with the progress of deep learning research, 
\eg~auto-differentiation, better optimization techniques, large-scale datasets, and fast computation devices, 
it is possible and in fact very easy to directly optimize a close approximation to AP.

%% file: arxiv_sec/background.tex
In this section, we define the  notations used throughout the paper.

\subsubsection{Task definition.} 
Given an input query, 
the goal of a retrieval system is to rank all instances in a retrieval set $\Omega = \{I_{i}, i = 0, \cdots, m\}$ based on their relevance to the query. 
For \emph{each} query instance $I_q$, 
the retrieval set is split into the positive $\mathcal{P}_q$ and negative $\mathcal{N}_q$ sets, 
which are formed by all instances of the same class and of different classes, respectively. 
Note that there is a different positive and negative set for each query.

\subsubsection{Average Precision~(AP).}
\label{sec:bg_ap}
AP is one of the standard metrics for information retrieval tasks~\cite{Manning08}. 
It is a single value defined as the area under a Precision-Recall curve. 
For a query $I_q$, 
the predicted relevance scores of all instances in the retrieval set are measured via a chosen metric. In our case, we use the cosine similarity (though the Smooth-AP method is independent of this choice):
\begin{align}
\mathcal{S}_{\Omega} = \left\{s_i = \left\langle \frac{v_q}{\|v_q\|} \cdot \frac{v_i}{\|v_i\|} \right\rangle, i = 0, \cdots, n\right\},
\end{align}
where $\mathcal{S}_{\Omega} = \mathcal{S}_P \cup \mathcal{S}_N$, 
and $\mathcal{S}_P = \{s_{\zeta}, \forall \zeta \in \mathcal{P}_q\}$, 
$\mathcal{S}_N = \{s_{\xi},   \forall \xi \in \mathcal{N}_q\}$
are the positive and negative relevance score sets, respectively, $v_{q}$ refers to the query vector, 
and $v_i$ to the vectorized retrieval set. 
The AP of a query $I_q$ can be computed as:
\begin{equation}  
\label{eq:AP_loss}
AP_{q} = \frac{1}{|\mathcal{S}_P|} \sum_{\substack{i \in \mathcal{S}_P}}
     \frac{\mathcal{R}(i, \mathcal{S}_P)}{\mathcal{R}(i, S_{\Omega})}, 
\end{equation}
where $\mathcal{R}(i,\mathcal{S}_P)$ and $\mathcal{R}(i,\mathcal{S}_\Omega)$ refer to the rankings of the instance $i$ in $\mathcal{P}$ and $\Omega$, respectively. 
Note that, the rankings referred to in this paper are assumed to be \textit{proper rankings}, 
meaning no two samples are ranked equally.

\subsubsection{Ranking Function~($\mathcal{R}$).}
\label{sub:interpreting}
Given that AP is a ranking-based method, 
the key element for direct optimisation is to define the ranking $\mathcal{R}$ of one instance $i$. 
Here, we define it in the following way~\cite{Qin10}:
\begin{align}
\mathcal{R}(i, \mathcal{S}) &= 1 +  
\sum_{\substack{j\in \mathcal{S}},j\neq i} \mathbb{1}\{(s_i - s_j) < 0\},
\end{align}
where $\mathbb{1}\{\cdot\}$ acts as an Indicator function, and $\mathcal{S}$ any set, \eg\ $\Omega$. 
Conveniently, this can be implemented by computing a difference matrix $D \in \mathbb{R}^{m \times m}$:
\begin{align}
D = \begin{bmatrix} 
    s_{1}    &  \dots   & s_{m} \\
    \vdots   &  \ddots  & \vdots \\ 
    s_{1}    &  \dots   & s_{m} 
    \end{bmatrix}
    - 
    \begin{bmatrix} 
    s_{1}   &  \dots   &   s_{1} \\
    \vdots  &  \ddots  &   \vdots \\
    s_{m}   &  \dots   &   s_{m} 
    \end{bmatrix}
\end{align}

\noindent 
The exact AP for a query instance $I_q$ from Eq.~\ref{eq:AP_loss} becomes:
\begin{align}  
\label{eq:AP}
AP_q 
= \frac{1}{|\mathcal{S}_P|} \sum_{\substack{i\in\mathcal{S}_P}} 
    \frac{1+ \sum_{j \in \mathcal{S}_p ,j\neq i} \mathbb{1}\{D_{ij} > 0\}}
       {1+ \sum_{j\in \mathcal{S}_P ,j\neq i} \mathbb{1}\{D_{ij} > 0\} + \sum_{j\in \mathcal{S}_N } \mathbb{1}\{D_{ij} > 0\}}
\end{align}

\subsubsection{Derivatives of Indicator.}
The particular Indicator function used in computing AP is a Heaviside step function $\mathcal{H}(\cdot)$~\cite{Qin10}, 
with its distributional derivative defined as Dirac delta function:
\begingroup\abovedisplayskip=6pt \belowdisplayskip=6pt
\begin{align*}
    \diff{\mathcal{H}(x)}{x} = \delta(x), 
\end{align*}
\endgroup
This is either flat everywhere, with zero gradient, or discontinuous, 
and hence cannot be optimized with gradient based methods (Figure~\ref{fig:sigmoid}).

%% file: arxiv_sec/method.tex
As explained, AP and similar metrics include a discrete ranking
function that is neither differentiable nor decomposable. 
In this section, we first describe {\em Smooth-AP}, 
which essentially replaces the discrete indicator function with a sigmoid function,
and then we provide an analysis on its relation to other 
ranking losses, such as triplet loss~\cite{HermansBeyer17Arxiv,Weinberger06},
FastAP~\cite{Cakir19} and Blackbox AP~\cite{Rolnek20optimizing}.

\subsection{Smoothing AP} 
\label{SmoothAPMethod}

\begin{figure}[t] 
\begin{center}
   \includegraphics[width=\linewidth]{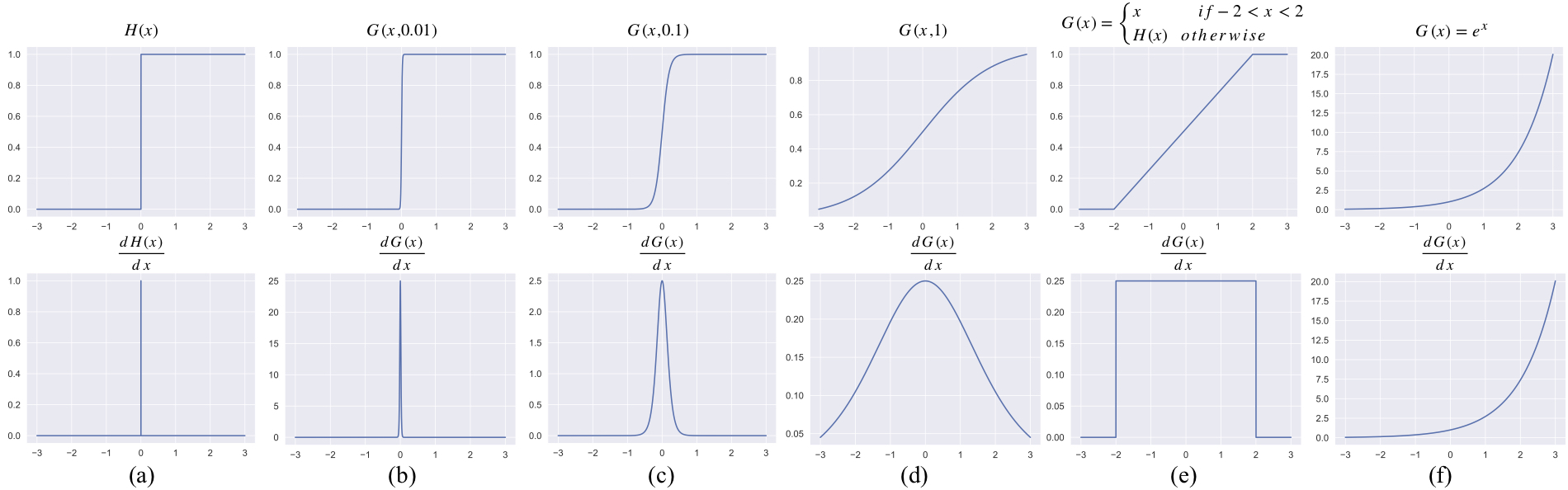}
\end{center}
\caption{\small{The possible \textbf{different approximations} to the discrete Indicator function. First row: Indicator function~(a), three sigmoids with increasing temperatures (b, c, d), linear~(e), exponential~(f). Second row: their derivatives.}}
\label{fig:sigmoid}
\end{figure}

To smooth the ranking procedure,  which will enable direct optimization of AP, Smooth-AP takes
a simple solution which is to replace the Indicator function $\mathbb{1}\{\cdot\}$ by a sigmoid function $\mathcal{G}(\cdot; \tau)$, where the $\tau$ refers to the temperature adjusting the sharpness:
\begin{align}
\mathcal{G}(x; \tau) = \frac{1}{1 + e^{\frac{-x}{\tau}}}.
\end{align}
Substituting $\mathcal{G}(\cdot; \tau)$ into Eq.~\ref{eq:AP}, the true AP can be approximated as:
\begin{align*}  
\label{smooth_AP_loss}
\footnotesize{AP_q \approx 
  \frac{1}{|\mathcal{S}_P|} \sum_{\substack{i\in\mathcal{S}_P}} 
  \frac{1 + \sum_{j \in \mathcal{S}_P} \mathcal{G}(D_{ij}; \tau)}
       {1 + \sum_{j\in \mathcal{S}_P} \mathcal{G}(D_{ij}; \tau) 
          + \sum_{j\in \mathcal{S}_N} \mathcal{G}(D_{ij}; \tau)}}
\end{align*}
\noindent with tighter approximation and convergence to the indicator function as $\tau \rightarrow 0$.
The objective function during optimization is denoted as:
\begingroup\abovedisplayskip=7pt \belowdisplayskip=7pt
\begin{align}
\mathcal{L}_{AP}  = \frac{1}{m}\sum^m_{k = 1} (1 - AP_k)
\end{align}
\endgroup

\noindent {\bf Smoothing parameter $\tau$ } 
governs the temperature of the sigmoid that replaces the Indicator function $\mathbb{1}\{\cdot\}$. 
It defines an operating region, where terms of the difference matrix are given a gradient by the Smooth-AP loss. 
If the terms are mis-ranked, Smooth-AP will attempt to shift them to the correct order. 
Specifically, a small value of $\tau$ results in a small operating region (Figure \ref{fig:sigmoid}~(b) -- note the small region with gradient seen in the sigmoid derivative), and a tighter approximation of true AP. 
The strong acceleration in gradient around the zero point (Figure \ref{fig:sigmoid}~(b)-(c) second row) is essential to replicating the desired qualities of AP, 
as it encourages the shifting of instances in the embedding space that result in a change of rank (and hence change in AP), 
rather than shifting instances by some large distance but not changing the rank. 
A large value of $\tau$ offers a large operating region, 
however, at the cost of a looser approximation to AP due to its divergence from the indicator function.\\

\noindent {\bf Relation to Triplet Loss.} 
\label{relation_triplet}
Here, we demonstrate that the triplet loss (a popular surrogate loss for ranking) is in fact optimising a distance metric rather than a ranking metric, which is sub-optimal when evaluating using a ranking metric.
As shown in Eq.~\ref{eq:AP}, 
the goal of optimizing AP is equivalent to minimizing all the $\sum_{i \in \mathcal{S}_P, j \in \mathcal{S}_N}\mathbb{1}\{D_{ij } < 0\} $, \ie~the violating terms. 
We term these as such because these terms refer to cases 
where a negative instance is ranked above a positive instance in terms of relevance to the query, 
and optimal AP is only acquired when all positive instances are ranked above all negative instances. 

For example, 
consider one query instance with predicted relevance score and ground-truth relevance labels as:
\begin{align*}
\text{Instances ordered by score}&:  (\mathit{s_{0}} \ \mathit{s_{4}} \  \mathit{s_{1}} \ \mathit{s_{2}} \ \mathit{s_{5}} \ \mathit{s_{6}} \ \mathit{s_{7}} \ \mathit{s_{3}}) \\
\text{Ground truth labels}&:  (1 \ \  0  \ \  1 \ \  1 \ \ 0 \ \ 0  \ \  0 \ \ 1)
\label{tab:equation_triplet}
\end{align*}

\noindent the violating terms are: 
$\{(s_4 - s_1), (s_4 - s_2), (s_4 - s_3), (s_5 - s_3), (s_6 - s_3), (s_7 - s_3)\}$.
An ideal AP loss would actually treat each of the terms unequally, 
\ie~the model would be forced to spend more capacity on shifting orders between $s_4$ and $s_1$, 
rather than $s_3$ and $s_7$, as that makes a larger impact on improving the AP.\\

Another interpretation of these violating cases can also be drawn from the triplet loss perspective.
Specifically, if we treat the query instance as an ``anchor'', with
$s_j$ denoting the similarity between the ``anchor'' and negative instance,
and $s_i$ denoting the similarity between the ``anchor'' and positive instance.
In this example, the triplet loss tries to optimize a margin hinge loss: 
\begin{align*}
\mathcal{L}_{\text{triplet}} & = \max(s_4 - s_1 + \alpha, 0) + \max(s_4 - s_2 + \alpha, 0) \\
& + \max(s_4 - s_3 + \alpha, 0) + \max(s_5 - s_3 + \alpha, 0) \\
& + \max(s_6 - s_3 + \alpha, 0)  + \max(s_7 - s_3 + \alpha, 0) 
\end{align*}
This can be viewed as a differentiable approximation to the goal of optimizing AP where the Indicator function has been replaced with the margin hinge loss, thus solving the gradient problem.
Nevertheless, 
using a triplet loss to approximate AP may suffer from two problems:
\emph{First},
all terms are linearly combined and treated equally in $\mathcal{L}_{\text{triplet}}$. 
Such a surrogate loss may force the model to optimize the terms that have only a small effect on  AP,
\eg\ optimizing $s_4 - s_1$ is the same as $s_7 - s_4$ in the triplet loss, 
however, from an AP perspective, 
it is important to correct the mis-ordered instances at high rank.
\emph{Second}, 
the linear derivative means that the optimization process is  purely based on distance (not ranking orders), 
which makes it sub-optimal when evaluating AP. 
For instance, in the triplet loss case, 
reducing the distance $s_4 - s_1$ from $0.8$ to $0.5$ is the same as from $0.2$ to $-0.1$. 
In practise, however, the latter case (shifting orders) will clearly have a much larger 
impact on the AP computation than the former.\\

\noindent {\bf Comparison to other AP-optimising methods.}
The two key differences between Smooth-AP 
and the recently introduced FastAP and Blackbox AP,
are that Smooth-AP (i) provides a closer approximation to Average Precision, 
and (ii) is far simpler to implement. 
Firstly, due to the sigmoid function, Smooth-AP optimises a ranking metric, 
and so has the same objective as Average Precision.
In contrast, 
FastAP and Blackbox AP linearly interpolate the non-differentiable~(piecewise constant) function,
which can potentially lead to the same issues as triplet loss, \ie~optimizing a distance metric, rather than rankings.
Secondly, Smooth-AP simply needs to replace the indicator function in the AP objective with a sigmoid function. 
While FastAP uses abstractions such as Histogram Binning, and Blackbox AP uses a variant of numerical derivative.
These differences are positively affirmed through the improved performance of Smooth-AP over several datasets~(Section~\ref{sec:experimental_results}).

%% file: arxiv_sec/experiment.tex
In this section, 
we describe the datasets used for evaluation, 
the test protocols,  and the implementation details. 
The procedure followed here is to take a pre-trained network and fine-tune with Smooth-AP loss. 
Specifically, 
ImageNet pretrained networks are used for the object/animal retrieval datasets, 
and high-performing face-verification models for the face retrieval datasets.

\subsection{Datasets}
We evaluate the  Smooth-AP loss on five datasets containing a wide range of domains and sizes. 
These include the commonly used retrieval benchmark datasets, as well as several additional
\textit{large-scale} ($>$100K images) datasets.  
Table~\ref{table:datasets} describes their details. \\

\setlength{\tabcolsep}{4pt}
\begin{table}[t]
\centering
\caption{\small{\textbf{Datasets} used for training and evaluation.}}

\begin{tabular}{|cl|r|r|r|}
\hline
\multicolumn{2}{|c|}{dataset}                                                                                                & \# Images & \# Classes & \# Ims/Class       \\ \hline
\multicolumn{1}{|c|}{}                                                                                              & SOP train                                              & 59,551    & 11,318     & 5.3          \\
\multicolumn{1}{|c|}{}                                                                                              & SOP test                                               & 60,502    & 11,316     & 5.3     \\
\multicolumn{1}{|c|}{}                                                                                              & VehicleID train                                              & 110,178    & 13,134     & 8.4         \\
\multicolumn{1}{|c|}{}                                                                                              & VehicleID test                                               & 40,365    & 4,800     & 8.4     \\
\multicolumn{1}{|c|}{}                                                                                              & INaturalist train                                      & 325,846   & 5,690      & 57.3        \\
\multicolumn{1}{|c|}{\multirow{-8}{*}{\begin{tabular}[c]{@{}c@{}}object/animal \\ retrieval datasets\end{tabular}}} & INaturalist test                                       & 136,093   & 2,452      & 55.5        \\ \cline{1-5}
\multicolumn{1}{|c|}{}                                                                                              & VGGFace2 train                                         & 3.31 M    & 8,631      & 363.7      \\
\multicolumn{1}{|c|}{}                                                                                              & VGGFace2 test                                          & 169,396   & 500        & 338.8       \\
\multicolumn{1}{|c|}{\multirow{-3}{*}{\begin{tabular}[c]{@{}c@{}}face retrieval \\ datasets\end{tabular}}}          & IJB-C                                                & 148,824         & 3,531          & 42.1         \\ \hline
\end{tabular}
\label{table:datasets}
\end{table}
\setlength{\tabcolsep}{1.4pt}

\noindent {\bf Stanford Online Product~(SOP)~\cite{Song16}} 
was initially collected for investigating the problem of metric learning.
It includes $120$K images of products that were sold online. 
We use the same evaluation protocol and train/test split as~\cite{Wang19ranked}. \\

\noindent {\bf VehicleID~\cite{Wah11}} 
contains $221,736$ images of $26,267$ vehicle categories, $13,134$ of which are used for training (containing $110,178$ images). By following the same test protocol as ~\cite{Wah11}, three test sets of increasing size are used for evaluation (termed small, medium, large), which contain $800$ classes ($7,332$ images), $1600$ classes ($12,995$ images) and $2400$ classes ($20,038$ images) respectively.\\

\noindent {\bf INaturalist~\cite{Van18}}
is a large-scale animal and plant species classification dataset, designed to replicate real-world scenarios through 461,939 images from 8,142 classes. 
It features many visually similar species, captured in a wide variety of environments. 
We construct a new image retrieval task from this dataset, 
by keeping 5,690 classes for training, 
and 2,452 unseen classes for evaluating image retrieval at test time,  according to the same test protocols as existing benchmarks~\cite{Wang19ranked}. 
We will make the train/test splits publicly available.\\

\noindent {\bf VGGFace2~\cite{Cao18}}
is a large-scale face dataset with over $3.31$ million images of $9,131$ subjects. 
The images have large variations in pose, age, illumination, ethnicity and profession, \eg\ actors, athletes, politicians.
For training, we use the pre-defined \emph{training set} with $8,631$ identities,
and for testing we use the \emph{test set} with $500$ identities, totalling $169$K testing images.\\

\noindent {\bf IJB-C~\cite{Maze18}}
is a challenging public benchmark for face recognition, containing images of subjects from both still frames and videos. Each video is treated as a single instance by averaging the CNN-produced vectors for each frame to a single vector. Identities with less than $5$ instances~(images or videos) are removed.

\subsection{Test Protocol}
Here, we describe the protocols for evaluating retrieval performance, 
mean Average Precision~(mAP) and  Recall@K~(R@K).
For all datasets, every instance of each class is used in turn as the query \(\mathit{I_q}\), 
and the retrieval set $\Omega$ is formed out of all the remaining instances. 
We ensure that each class in all datasets contains several images (Table~\ref{table:datasets}), 
such that if an instance from a class is used as the query, 
there are plenty of remaining positive instances in the retrieval set. 
For object/animal retrieval evaluation, 
we use the \textit{Recall@K} metric in order to compare to existing works. 
For face retrieval, AP is computed from the resulting output ranking for each query,
and the mAP score is computed by averaging the APs across every instance in the dataset, resulting in a single value. 

\subsection{Implementation Details}
\noindent {\bf Object/animal retrieval~(SOP, VehicleID, INaturalist).}
In line with previous works~\cite{Cakir19,Rolnek20optimizing,roth2019mic,sanakoyeu2019divide,wang2020cross,Wu17},we use ResNet50~\cite{He16} as the backbone architecture, 
which was pretrained on ImageNet~\cite{imagenet}.
We replace the final softmax layer with one linear layer~(following~\cite{Cakir19,Rolnek20optimizing}, 
with dimension being set to $512$).
All images are resized to $256\times256$. 
At training time, we use random crops and flips as augmentations, 
and at test time, a single centre crop of size $224 \times 224$ is used.
For all experiments we set $\tau$ to 0.01~(Section~\ref{ablation}).\\

\noindent {\bf Face retrieval datasets~(VGGFace2, IJB-C).}
We use two high performing face verification networks: the method from~\cite{Cao18} using the SENet-50 architecture~\cite{Hu2018} and the state-of-the-art ArcFace~\cite{Deng19} (using ResNet-50), 
both trained on the VGGFace2 training set. 
For SENet-50, 
we follow~\cite{Cao18} and use the same face crops (extended by the recommended amount), 
resized to $224 \times 224$ and we L2-normalize the final 256D embedding.
For ArcFace, we generate normalised face crops $(112 \times 112)$ by using the provided face detector~\cite{Deng19}, 
and align them with the predicted $5$ facial key points, 
then L2-normalize the final $512$D embedding. 
For both models, we set the batch size to 224 and $\tau$ to 0.01 (Section~\ref{ablation}). \\

\noindent {\bf Mini-batch training.}
During training, we form each mini-batch by randomly sampling classes such that each represented class has $|\mathcal{P}|$ samples per class. For all experiments, we L2-normalize the embeddings, use cosine similarity to compute the relevance scores between the query and the retrieval set, set $|\mathcal{P}|$ to 4, 
and use an Adam~\cite{KingmaB14} optimiser with a base learning rate of $10^{-5}$ with weight decay $4e^{-5}$.
We employ the same hard negative mining technique as~\cite{Cakir19,Rolnek20optimizing} only for the Online Products dataset. 
Otherwise we use no special sampling strategies.

\begin{figure*}[t!]
\begin{center}
\includegraphics[width=\linewidth]{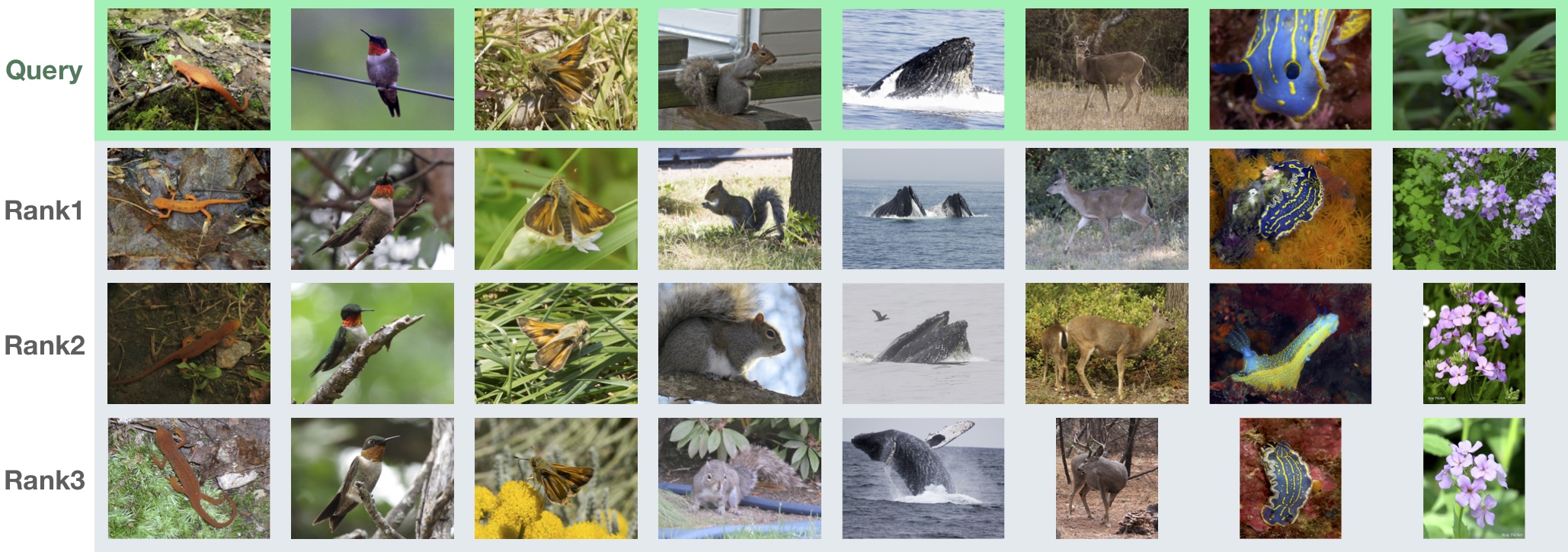}
\end{center}
\caption{\small{
\textbf{Qualitative results for the INaturalist dataset using Smooth-AP loss}. For each query image (top row), the top 3 instances from the retrieval set are shown ranked from top to bottom. Every retrieved instance shown is a true positive. }}
\label{fig:short}
\end{figure*}

%% file: arxiv_sec/results.tex
In this section, 
we first explore the effectiveness of the proposed Smooth-AP by examining the performance of various models on the five retrieval datasets.
Specifically, 
we compare with the recent AP optimization and broader metric learning methods on the standard benchmarks
SOP and VehicleID (Section~\ref{sub:eval_sopcub}), 
and then shift to further large-scale experiments, 
\eg~INaturalist for animal/plant retrieval, 
and IJB-C and VGGFace2 for face retrieval (Sections~\ref{sub:eval_inatur}-\ref{sub:eval_face}). 
Then, we present an ablation study of various
hyper-parameters that affect the performance 
of Smooth-AP: 
the sigmoid temperature, the size of the positive set, and the batch size
(Section~\ref{ablation}).
Finally, we discuss various findings and analyze the performance gaps between various models (Section~\ref{sub:discussion}). \\

Note that, although there has been a rich literature on metric learning methods \cite{Duan18deep,Ge18,Harwood17smart,Kim18,Law17,Lu19,Movshovitz17,oh2017deep,Opitz17,qian2019softtriple,Rolnek20optimizing,Song16,suh2019stochastic,Ustinova16,Wang17metric,Wang19ranked,wang2019multi,wang2020cross,Wu17,Xuan18deep,Yuan17hard} using these image retrieval benchmarks,
we only list the very recent state-of-the-art approaches, 
and try to compare with them as fairly as we can, \eg\ no model ensemble, and using the same backbone network and image resolution. 
However, there still remain differences on some small experimental details, 
such as embedding dimensions, optimizer, and learning rates.
Qualitative results for the INaturalist dataset are illustrated in Figure~\ref{fig:short}.

\subsection{Evaluation on Stanford Online Products~(SOP)}
\label{sub:eval_sopcub}
\begin{figure}[!t]
\begin{minipage}{\textwidth}
  \begin{minipage}[b]{0.55\textwidth}
    \centering
    \footnotesize
    \begin{tabular}{l|cccc}
                          & \multicolumn{4}{c}{\textbf{SOP}}                  \\ \hline
\textit{Recall@K}         & 1             & 10            & 100           & 1000          \\ \hline
Margin  ~\cite{Wu17}                        & 72.7          & 86.2          & 93.8          & 98.0          \\
Divide~\cite{sanakoyeu2019divide}                        & 75.9          & 88.4          & 94.9          & 98.1          \\
FastAP~\cite{Cakir19}                     & 76.4          & 89.0          & 95.1          & 98.2          \\
MIC~\cite{roth2019mic}                       & 77.2          & 89.4          & 95.6          & -             \\
Blackbox AP~\cite{Rolnek20optimizing}               & 78.6          & 90.5          & 96.0          & 98.7          \\
Cont. w/M~\cite{wang2020cross}                 & \textbf{80.6} & \textbf{91.6} & 96.2          & 98.7          \\ \hline
\textbf{Smooth-AP BS=224} & 79.2          & 91.0          & 96.5          & 98.9          \\
\textbf{Smooth-AP BS=384} & 80.1          & 91.5          & \textbf{96.6} & \textbf{99.0}
\end{tabular}
      \captionof{table}{\small{\textbf{Results on Stanford Online Products.} 
Deep metric learning and recent AP approximating methods are compared to using the ResNet50 architecture. 
BS: mini-batch size.
}}
\label{tab:object_ex}
\end{minipage}
\hfill
   \begin{minipage}[b]{0.43\textwidth}
    \centering
	\includegraphics[width=.96\textwidth]{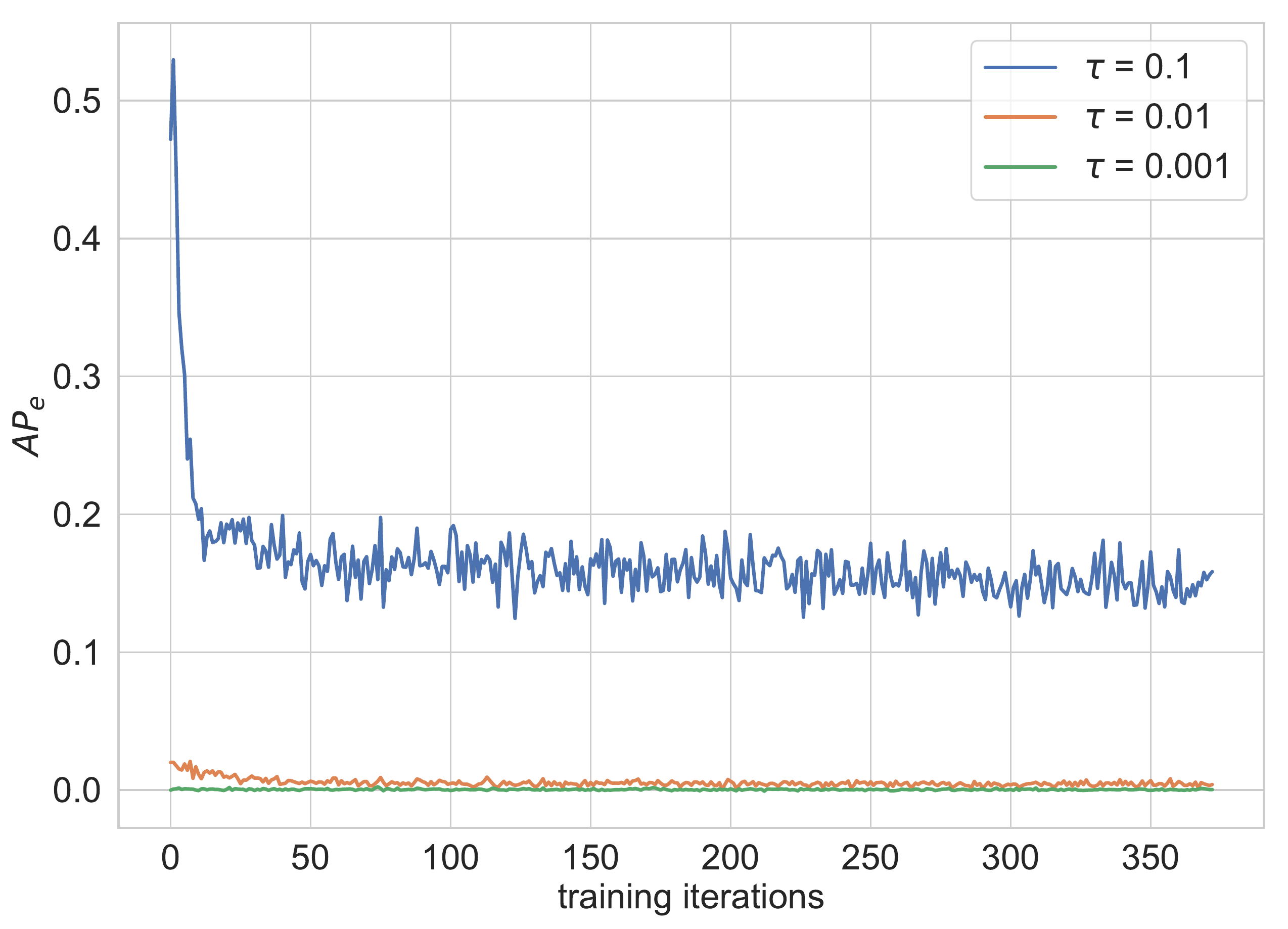}
    \captionof{figure}{{\small{
    {\bf The AP approximation error}, $AP_e$ over one training epoch for Online Products for different values of sigmoid annealing temperature, $\tau$. }}}
    \label{approx_fig}
   \end{minipage}
\end{minipage}
\end{figure}

We compare with a wide variety of state-of-the-art image retrieval methods, 
\eg~deep metric learning methods~\cite{roth2019mic,sanakoyeu2019divide,wang2020cross,Wu17}, 
and AP approximation methods~\cite{Cakir19,Rolnek20optimizing}.
As shown in Table~\ref{tab:object_ex}, 
we observe that Smooth-AP achieves state-of-the-art results on the SOP benchmark. 
In particular, our best model outperforms the very recent AP approximating methods~(Blackbox~AP and FastAP) by a $1.5\%$ margin for Recall@1. 
Furthermore, Smooth-AP performs on par with the concurrent work~(Cross-Batch Memory~\cite{wang2020cross}).
This is particularly impressive as~\cite{wang2020cross} harnesses memory techniques to sample from many mini-batches simultaneously for each weight update, 
whereas Smooth-AP only makes use of a single mini-batch on each training iteration. \\

Figure~\ref{approx_fig} provides a quantitative analysis into the effect of sigmoid temperature $\tau$ on the tightness of the AP approximation, 
which can be plotted via the AP approximation error:
\begin{equation}
AP_e = |AP_{pred} - AP|
\end{equation}
where $AP_{pred}$ is the predicted approximate AP when the sigmoid is used in place of the indicator function in Equation 5, 
and $AP$ is the true AP. 
As expected, a lower value of $\tau$ leads to a tighter approximation to Average Precision, 
shown by the low approximation error.

\setlength{\tabcolsep}{0.5pt}

\subsection{Evaluation on VehicleID and INaturalist}
\label{sub:eval_inatur}

In Table~\ref{tab:Inat}, we show results on the VehicleID and INaturalist dataset. 
We observe that Smooth-AP achieves state-of-the-art results on the challenging and large-scale VehicleID dataset. In particular, our model outperforms FastAP by a significant $3\%$ for the Small protocol Recall@1. 
Furthermore, Smooth-AP exceeds the performance of~\cite{wang2020cross} on 4 of the 6 recall metrics. 

As we are the first to report results on INaturalist for {\em image retrieval}, 
in addition to Smooth-AP, 
we re-train state-of-the-art metric learning and AP approximating methods, 
with the respective official code,
\eg~Triplet and ProxyNCA~\cite{triplet_nca}, FastAP~\cite{fast_ap}, Blackbox~AP~\cite{blackbox_ap}.
As shown in Table~\ref{tab:Inat}.
Smooth-AP outperforms all methods by $2-5\%$ on Recall@1 for the experiments when the same batch size is used (224). Increasing the batch size to 384 for Smooth-AP leads to a further boost of 1.4\% to 66.6 for Recall@1. These results demonstrate that Smooth-AP is particularly suitable for {\em large-scale} retrieval datasets, 
thus revealing its scalability to {\em real-world} retrieval problems. We note here that these large-scale datasets ($>$100k images) are less influenced by hyper-parameter tuning and so provide ideal test environments to demonstrate improved image retrieval techniques. 

\setlength{\tabcolsep}{2pt}
\begin{table}[t]	
\centering
\footnotesize
\caption{\small{\textbf{Results on the VehicleID~(left) and INaturalist~(right).} All experiments are conducted using ResNet50 as backbone. All results for INaturalist are from publicly available official implementations in the PyTorch framework with a batch size of 224. 
$\dagger$ refers to the recent re-implementation~\cite{triplet_nca} - we make the design choice for Proxy NCA loss to keep the number of proxies equal to the number of training classes. The VehicleID results are obtained with a batch-size of 384.}}
\resizebox{\textwidth}{!}{%
	\begin{tabular}{l|cccccc}
\hline
\textit{}                                               & \multicolumn{6}{c}{\textbf{VehicleID}}                                                            \\ \hline
\multicolumn{1}{c|}{}                                   & \multicolumn{2}{c}{Small}    & \multicolumn{2}{c}{Medium}    & \multicolumn{2}{c}{Large} \\
Recall@K                                                & 1             & 5             & 1             & 5             & 1           & 5           \\ \hline
Divide~\cite{sanakoyeu2019divide} & 87.7          & 92.9          & 85.7          & 90.4          & 82.9        & 90.2        \\
MIC~\cite{roth2019mic}            & 86.9          & 93.4          & -             & -             & 82.0        & 91.0        \\
FastAP~\cite{Cakir19}             & 91.9          & 96.8          & 90.6          & 95.9          & 87.5        & 95.1        \\
Cont. w/M~\cite{wang2020cross}    & 94.7          & 96.8          & \textbf{93.7} & 95.8          & \textbf{93.0}        & 95.8        \\ \hline
\textbf{Smooth-AP}                                      & \textbf{94.9} & \textbf{97.6} & 93.3          & \textbf{96.4} & 91.9        & \textbf{96.2}        \\ \hline
\end{tabular}
\quad
	\begin{tabular}{l|cccc}
\hline
\textit{}                                                    & \multicolumn{4}{c}{\textbf{INaturalist}} \\ \hline
\textit{Recall@K}                                            & 1        & 4        & 16       & 32       \\ \hline
Triplet Semi-Hard~\cite{Wu17}          & 58.1     & 75.5     & 86.8     & 90.7     \\
Proxy NCA$\dagger$~\cite{Movshovitz17} & 61.6     & 77.4     & 87.0     & 90.6     \\
FastAP~\cite{Cakir19}                  & 60.6     & 77.0     & 87.2     & 90.6     \\
Blackbox AP~\cite{Rolnek20optimizing}  & 62.9     & 79.0     & 88.9     & 92.1     \\ \hline
\textbf{Smooth-AP BS=224}          & 65.9     & 80.9     & 89.8     & 92.7     \\
\textbf{Smooth-AP BS=384}          & \textbf{67.2}        & \textbf{81.8}        & \textbf{90.3}        & \textbf{93.1}        \\ \hline
\end{tabular} %

\label{tab:Inat}
}
\end{table}

\subsection{Evaluation on Face Retrieval} 
\label{sub:eval_face}

\setlength{\tabcolsep}{4pt}
\begin{table}[t]
\centering
\footnotesize
\caption{\small{\textbf{mAP results on face retrieval datasets}. 
Smooth-AP {\em consistently} boosts the AP performance for both VGGFace2 and ArcFace, 
while outperforming other standard metric learning losses (Pairwise contrastive and Triplet).}}
\label{tab:face_retrieval_results}
	\begin{tabular}{c|cc}
                        \hline
                        VGGFace2 & VF2 Test & IJB-C \\ \hline
                        Softmax                &  0.828   & 0.726 \\
                        +Pairwise              &  0.828   & 0.728 \\
                        +Triplet             &  0.845   & 0.740 \\ 
                       \textbf{+Smooth-AP}             &  {\bf 0.850}   & {\bf 0.754} \\ \hline
	\end{tabular}
\quad
	\begin{tabular}{c|cc}
                        \hline
                        ArcFace & VF2 Test & IJB-C \\ \hline
                        ArcFace            &  0.858   & 0.772 \\
                        +Pairwise               &  0.861   & 0.775 \\
                        +Triplet              &  0.880   & 0.787 \\ 
                  \textbf{+Smooth-AP}   & \textbf{0.902}    & \textbf{0.803} \\ \hline
	\end{tabular}
\end{table}

Due to impressive results~\cite{Cao18,Deng19}, face retrieval is considered saturated. Nevertheless, we demonstrate here that Smooth-AP can further boost the face retrieval performance. 
Specifically, 
we append Smooth-AP on top of modern methods (VGGFace2 and ArcFace) and evaluate mAP on IJB-C and 
VGGFace2, \ie\ one of the largest face recognition datasets. 

As shown in Table~\ref{tab:face_retrieval_results},
when appending the Smooth-AP loss,
retrieval metrics such as mAP can be significantly improved upon the baseline model for both datasets. 
This is particularity impressive as both baselines have already shown very strong performance on facial verification and identification tasks, 
yet Smooth-AP is able to increase mAP by up to $4.4\%$ on VGGFace2 and $3.1\%$ on ArcFace. 
Moreover, Smooth-AP strongly outperforms both the pairwise~\cite{Chopra05}  
and triplet~\cite{Schroff15} losses, 
\ie\ the two most popular surrogates to a ranking loss. 
As discussed in Section~\ref{SmoothAPMethod}, 
these surrogates optimise a distance metric rather than a ranking metric, 
and the results show that the latter is optimal for AP.

\subsection{Ablation study}
\label{ablation}

To investigate the effect of different hyper-parameter settings, 
\eg\ the sigmoid temperature $\tau$, the size of the positive set $|\mathcal{P}|$, 
and batch size $B$ (Table~\ref{ablation_table}),
we use VGGFace2 and IJB-C with SE-Net50~\cite{Cao18},
as the large-scale datasets are unlikely to lead to overfitting, 
and therefore provide a fair understanding about these hyper-parameters.
Note that we only vary one parameter at a time.\\

\input{table/ablation.tex}


\noindent {\bf Effect of sigmoid temperature $\tau$.}
As explained in Section~\ref{SmoothAPMethod}, 
$\tau$ governs the smoothing of the sigmoid that is used to approximate the indicator function in the Smooth-AP loss. The ablation shows that a value of $0.01$ leads to the best mAP scores, which is the optimal trade-off between AP approximation and a large enough operating region in which to provide gradients.
Surprisingly, this value~($0.01$) corresponds to a small operating region.
We conjecture that a tight approximation to true AP is the key, 
and when partnered with a large enough batch size, 
enough elements of the difference matrix will lie within the operating region in order to induce sufficient re-ranking gradients. 
The sigmoid temperature can further be viewed from the margin perspective (inter-class margins are commonly used in metric learning to help generalisation~\cite{Chopra05,Hyun16,Weinberger06}). 
Smooth-AP only stops providing gradients to push a positive instance above a negative instance once they are a distance equal to the width of the operating region apart, 
hence enforcing a margin that equates to roughly 0.1 for this choice of $\tau$. \\

\noindent {\bf Effect of positive set $|\mathcal{P}|$.}
In this setting, 
the positive set represents the \textit{instances} that come from the same class in the mini-batch during training.
We observe that a small value ($4$) results in the highest mAP scores, 
this is because mini-batches are formed by sampling at the class level, 
where a low value for $|\mathcal{P}|$ means a larger number of sampled classes and a higher probability of sampling hard-negative instances that violate the correct ranking order.
Increasing the number of classes in the batch results in a better batch approximation of the true class distribution, 
allowing each training iteration to enforce a more optimally structured embedding space. \\

\noindent {\bf Effect of batch size $B$.}
Table~\ref{ablation_table} shows that large batch sizes result in better mAP, especially for VGGFace2. 
This is expected, as it again increases the chance of getting hard-negative samples in the batch.

\subsection{Further discussion}
\label{sub:discussion}
There are several important observations in the above results. 
Smooth-AP outperforms all previous AP approximation approaches, 
as well as the metric learning techniques~(pair, triplet, and list-wise) on \emph{three} image retrieval benchmarks, 
SOP, VehicleID, Inaturalist,
with the performance gap being particularly apparent on the large-scale \textit{INaturalist} dataset.
Similarly, when scaled to face datasets containing millions of images, 
Smooth-AP is able to improve the retrieval metrics for state-of-the-art face verification networks.
We hypothesis that these performance gains upon the previous AP approximating methods come from a tighter approximation to AP than other existing approaches,
hence demonstrating the effectiveness and scalability of Smooth-AP.
Furthermore, many of the properties that deep metric learning losses handcraft into their respective methods (distance-based weighting~\cite{Song16,Wu17}, 
inter-class margins~\cite{Chopra05,Schroff15,Song16,Wang19ranked}, 
intra-class margins~\cite{Wang19ranked}), are naturally built into our AP formulation, 
and result in improved generalisation capabilities.

%% file: table/ablation.tex
\setlength{\tabcolsep}{4pt}
\begin{table}[t]
\caption{\small{{\bf Ablation study} over different parameters: ~temperature $\tau$, size of positive set during minibatch sampling $|\mathcal{P}|$, and batch size $B$. 
Performance is benchmarked on VGGFace2-Test and IJB-C.}}       
\label{ablation_table}
\centering
            \footnotesize
                    \begin{tabular}{c|cc}
                        \hline
                        $\tau$ & \multicolumn{2}{c}{mAP} \\ \hline
                        
\multicolumn{1}{c|}{} &  \multicolumn{1}{c}{VF2} & \multicolumn{1}{c}{IJB-C} \\ \hline
                        0.1                 &  0.824 & 0.726  \\
                        \textbf{0.01}            & \textbf{0.844} & \textbf{0.736}   \\
                        0.001                  & 0.839 & 0.733  \\ \hline
                        \multicolumn{3}{c}{\small{$|\mathcal{P}|$ = 4, $B=$ 128}} 
                    \end{tabular}
                \hfill
                \begin{tabular}{c|cc}
                        \hline
                        $|\mathcal{P}|$ & \multicolumn{2}{c}{mAP}  \\ \hline
                        \multicolumn{1}{c|}{} &  \multicolumn{1}{c}{VF2} & \multicolumn{1}{c}{IJB-C} \\ \hline
                        \textbf{4}                 &  \textbf{0.844} & \textbf{0.736}  \\
                        8                & 0.833 & 0.734   \\
                        16                  & 0.824 & 0.726  \\ \hline
                        \multicolumn{3}{c}{\small{$\tau$ = 0.01, $B=$ 128}}
                    \end{tabular}
                \hfill
                \begin{tabular}{c|cc}
                        \hline
                        $|\mathcal{B}|$ & \multicolumn{2}{c}{mAP}  \\ \hline
                        \multicolumn{1}{c|}{} &  \multicolumn{1}{c}{VF2} & \multicolumn{1}{c}{IJB-C} \\ \hline
                        64                 &  0.824  &  0.726  \\
                        128            & 0.844 & 0.736   \\
                        \textbf{256}.  & \textbf{0.853}  & \textbf{0.754}  \\ \hline
                        \multicolumn{3}{c}{\small{$\tau$ = 0.01, $|\mathcal{P}|$ = 4}}
                    \end{tabular}

\end{table}

%% file: arxiv_sec/conclusion.tex
We introduce \textit{Smooth-AP}, 
a novel loss that directly optimizes a smoothed approximation of AP. 
This is in contrast to modern contrastive, 
triplet, and list-wise deep metric learning losses which act as surrogates to encourage ranking.  
We show that Smooth-AP outperforms recent AP-optimising methods, 
as well as the deep metric learning methods, 
and with a simple and elegant, plug-and-play style method. 
We provide an analysis for the reasons why Smooth-AP outperforms these other losses, 
\ie~Smooth-AP preserves the goal of AP which is to optimise ranking rather than distances in the embedding space. 
Moreover, we also show that fine-tuning face-verification networks by appending the Smooth-AP loss 
can strongly improve the performance. 
Finally, in an effort to bridge the gap between experimental settings and real-world retrieval scenarios, 
we provide experiments on several large-scale datasets and 
show Smooth-AP loss to be considerably more scalable than previous approximations.

\subsubsection{Acknowledgements.}
We are grateful to Tengda Han, Olivia Wiles, Christian Rupprecht, Sagar Vaze, Quentin Pleple and Maya Gulieva for proof-reading, and to Ernesto Coto for the initial motivation for this work. 
Funding for this research is provided by the EPSRC Programme Grant Seebibyte EP/M013774/1. AB is funded by an EPSRC DTA Studentship. 

%% file: arxiv_sec/E.tex
\begin{figure}[htp]
\begin{center}
   \includegraphics[width=1\linewidth]{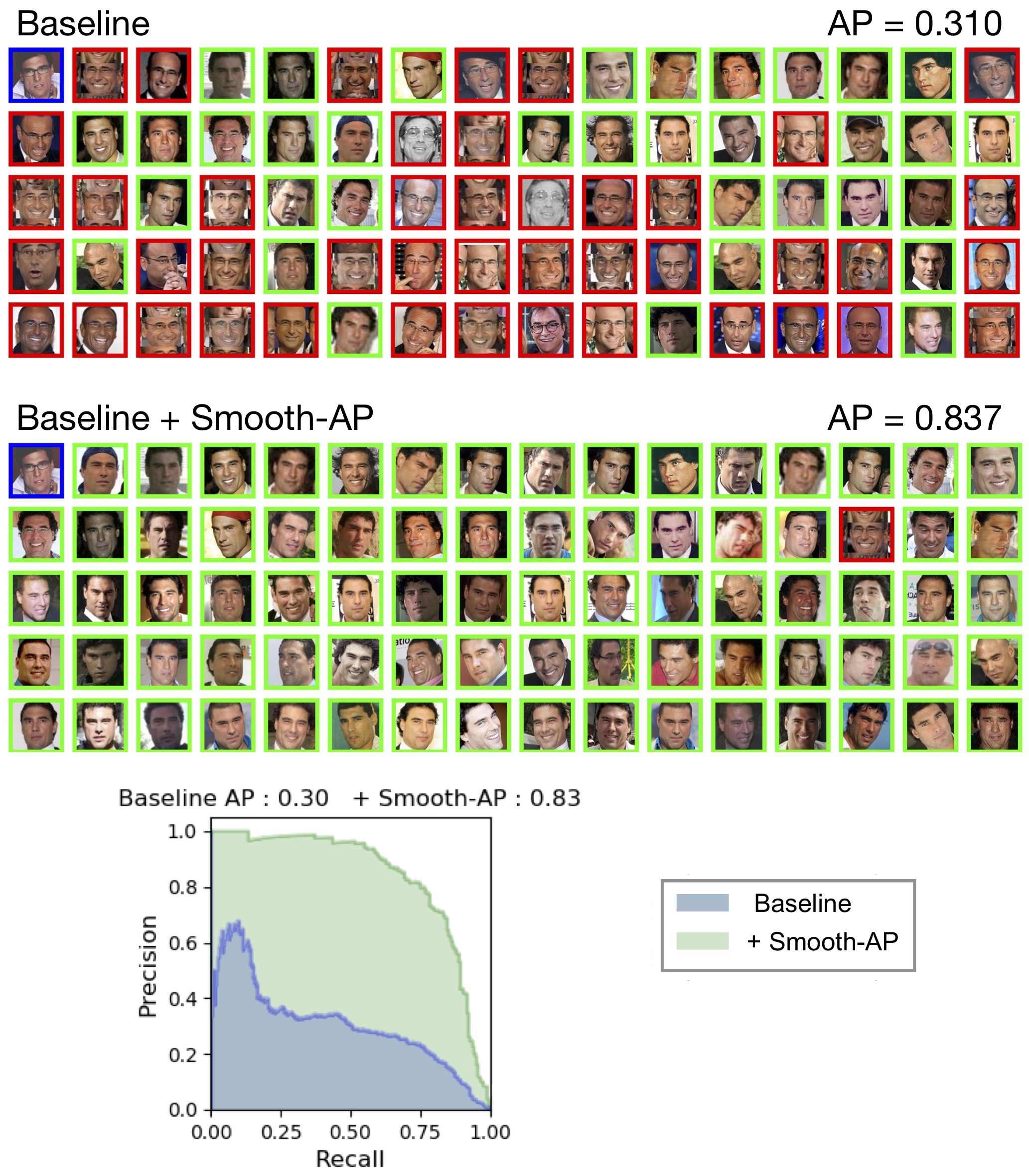}
\end{center}
\vspace{-10pt}
\beforecaptions
   \caption{\small{Qualitative results from the VGGFace2 Test set for the SENet-50~\cite{Cao18} baseline model. We show a query instance (blue) and the first 79 ranked instances in the retrieval set for the baseline model both before and after Smooth-AP was appended (ranked from left to right, starting next to the query). As shown by the precision-recall curves, Smooth-AP causes the Average Precision to jump by an impressive 52.9\%, and the number of false positives (shown in red) in the top ranked retrieved instances drops considerably. The size of the positive set for each instance in the VGGFace2 test set,  $|P|\approx338$.}}
 \aftercaptions
 \vspace{-1mm}
 \label{face_1}
\end{figure}
\begin{figure}[htp]
\begin{center}
   \includegraphics[width=1\linewidth]{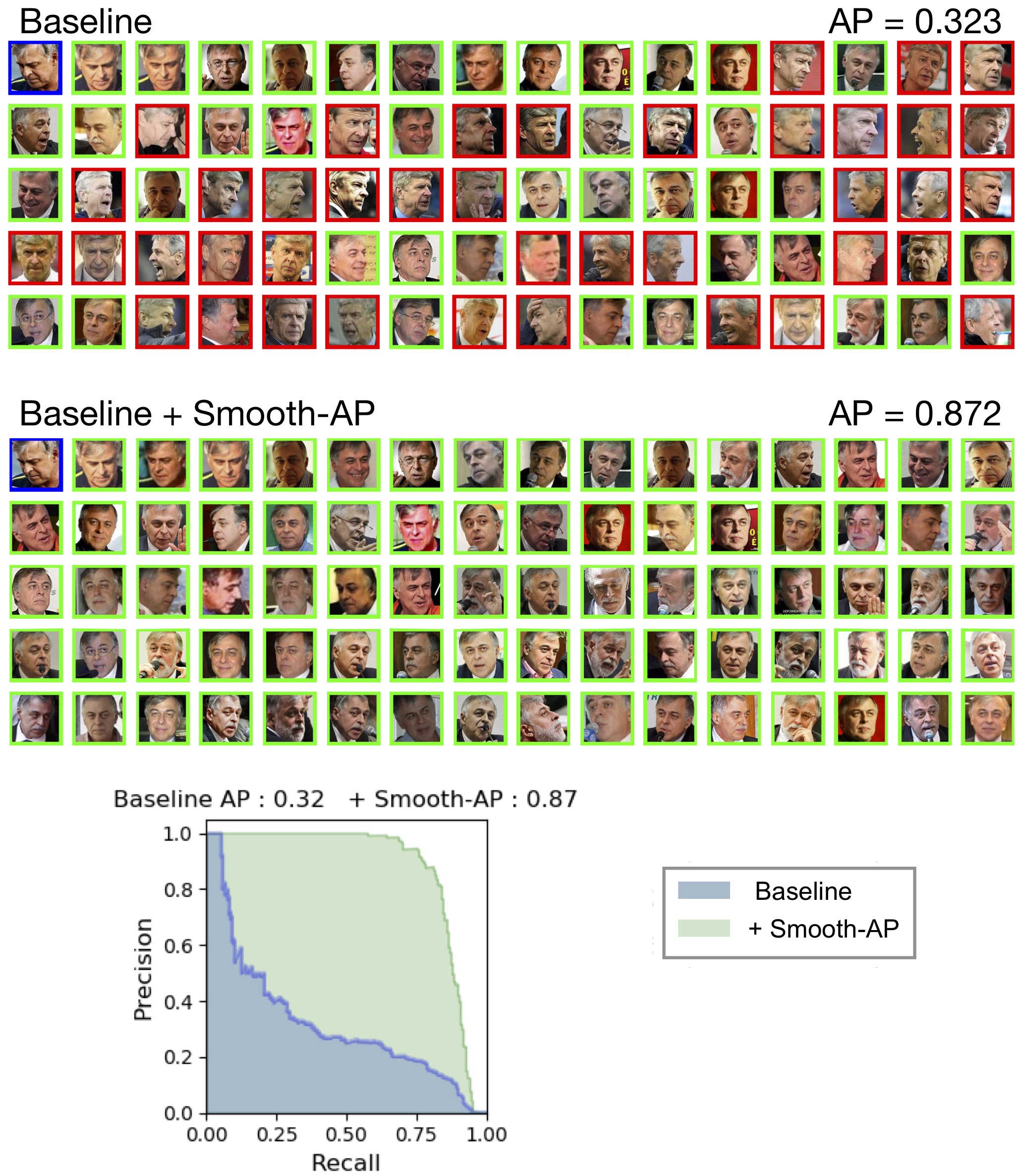}
\end{center}
\vspace{-10pt}
\beforecaptions

    \caption{\small{Here, we show the qualitative results for a second query instance for the SENet-50~\cite{Cao18} baseline model on the VGGFace2 test set, both before and after appending the Smooth-AP loss. Here, we see another large increase in Average Precision of 54.9\% caused by the addition of the Smooth-AP loss. We see that all false positives (shown in red) are removed from the top ranked retrieved instances after adding the Smooth-AP loss. Take the situation where each row of top-ranked instances corresponds to pages of retrieved results that a user is presented with when using a retrieval system. With the baseline model, the user comes across many false positives in the first few pages. After appending the Smooth-AP loss, the user encounters no false positives in at least the first five pages. This demonstrates the benefit to {\em user experience} in appending Smooth-AP to a retrieval network.}}

 \aftercaptions
 \vspace{-1mm}
 \label{face_2}
\end{figure}
\begin{figure}[htp]
\begin{center}
   \includegraphics[width=1\linewidth]{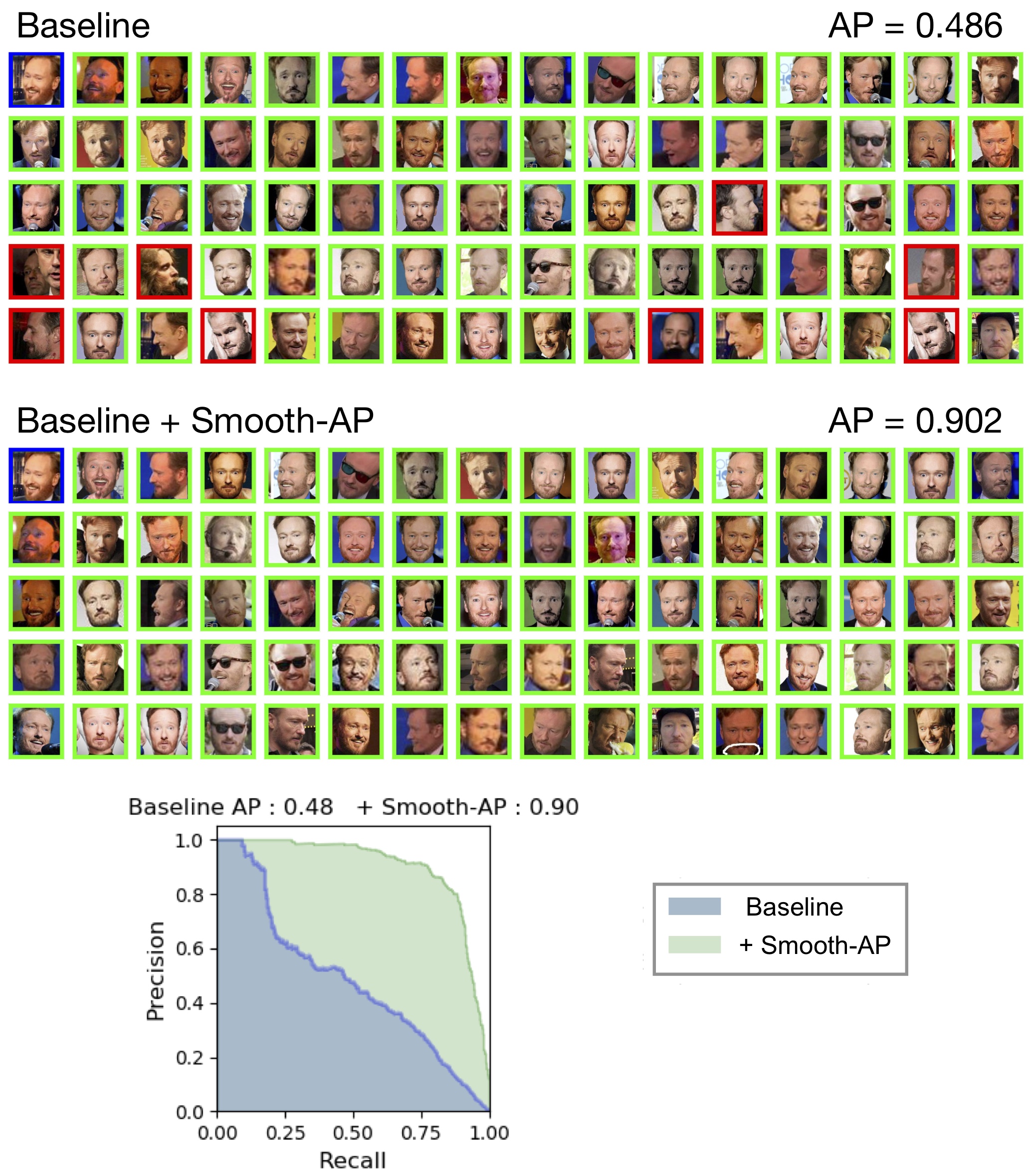}
\end{center}
\vspace{-10pt}
\beforecaptions
   \caption{\small{Here, we show the qualitative results for a third query instance for the SENet-50~\cite{Cao18} baseline model on the VGGFace2 test set, both before and after appending the Smooth-AP loss. We see a large improvement in Average Precision of 41.8\% after adding the Smooth-AP loss and the removal of all false positives from the top ranked retrieval results. These results confirm that Smooth-AP is indeed tailored to addressing the ranking issue.}}
 \aftercaptions
 \vspace{-1mm}
 \label{face_3}
\end{figure}
\begin{figure}[htp]
\begin{center}
   \includegraphics[width=1\linewidth]{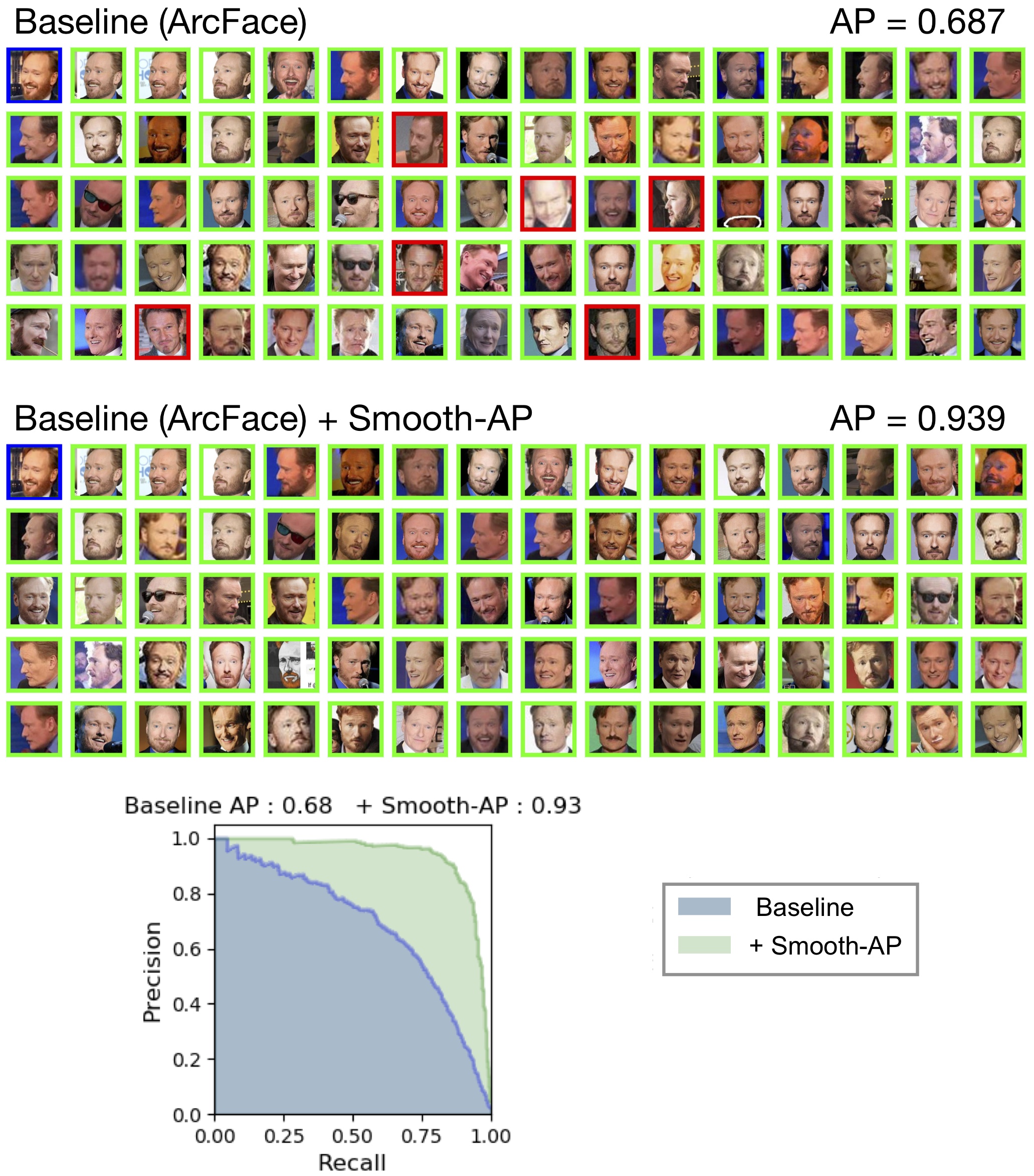}
\end{center}
\vspace{-10pt}
\beforecaptions
   \caption{\small{Here, we show the qualitative results for a query instance from the VGGFace2 test set (same as in Figure~\ref{face_3}) for the state-of-the-art ArcFace (ResNet-50)~\cite{Deng19} baseline, both before and after appending the Smooth-AP loss. Appending the Smooth-AP loss to this impressive baseline leads to a large gain in Average Precision (24.8\%), and again to the removal of all false positives from the top ranked retrieval results. This demonstrates that state-of-the-art face retrieval networks are far from saturated on the Average Precision metric, and through appending the simple Smooth-AP loss, this metric and the resulting user experience when using the face retrieval system can be greatly improved.}}
 \aftercaptions
 \vspace{-1mm}
 \label{face_4}
\end{figure}
\begin{figure}[htp]
\begin{center}
   \includegraphics[width=1\linewidth]{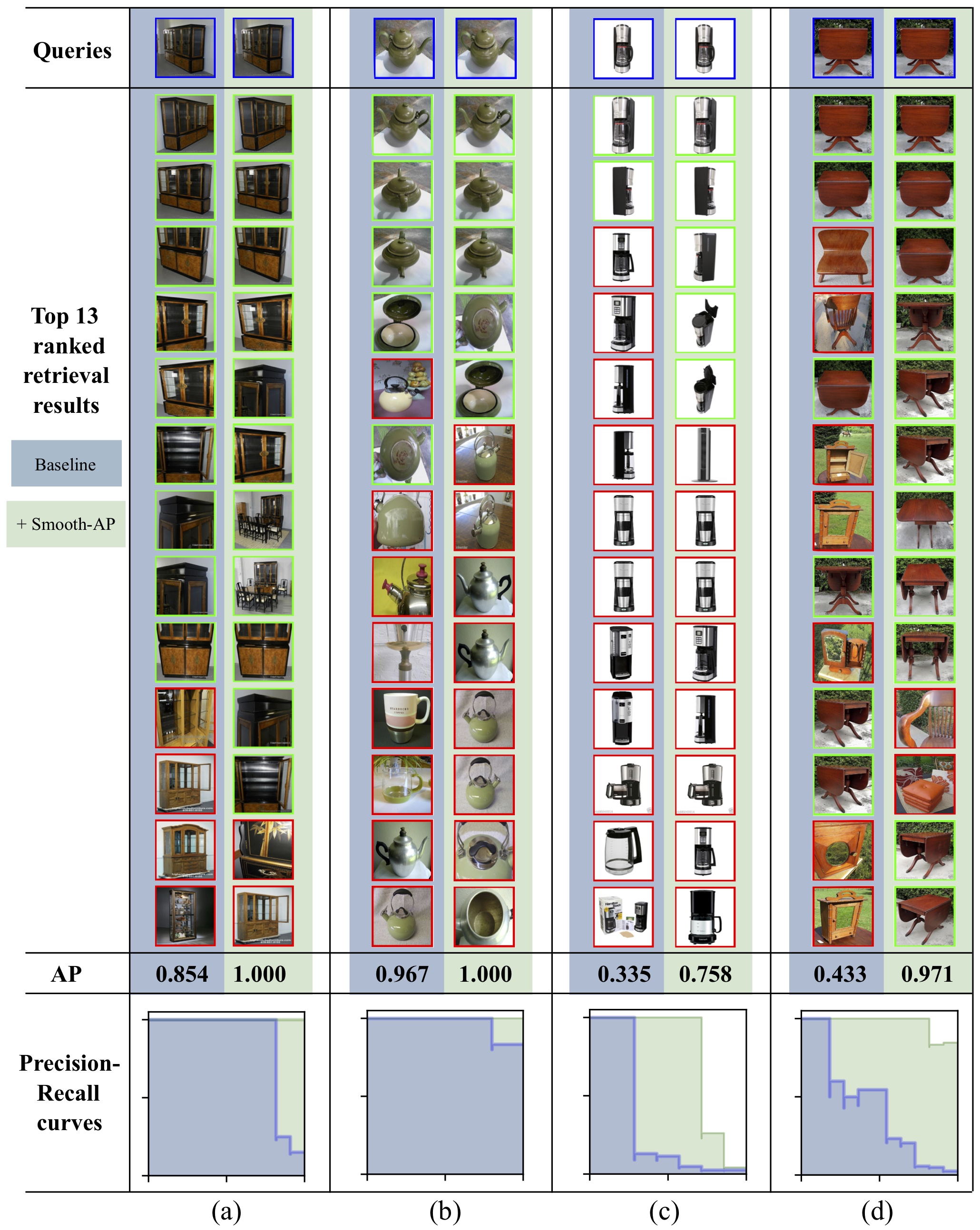}
\end{center}
\vspace{-10pt}
\beforecaptions
   \caption{\small{This Figure shows four separate query instances from the online products dataset and the top ranked instances from the retrieval set when using the baseline model (ImageNet pre-trained weights), and after appending the Smooth-AP loss. It is noted that for this dataset, the size of the positive set ($|P|$) in the retrieval set is very small ($|P|=$ 11, 5, 7, 11 for (a),(b),(c),(d) respectively), and so for cases (a) and (b) all positive instances are shown correctly retrieved above all false positives (also indicated by the AP=1.00) for the Smooth-AP model. Particularly \textit{impressive} are the examples (b) and (c), where instances from the positive set which depict a far different pose from the query are retrieved above false positives that are very visually similar to the query. }}
 \aftercaptions
 \vspace{-1mm}
 \label{SOP}
\end{figure}
\begin{figure}[htp]
\begin{center}
   \includegraphics[width=1\linewidth]{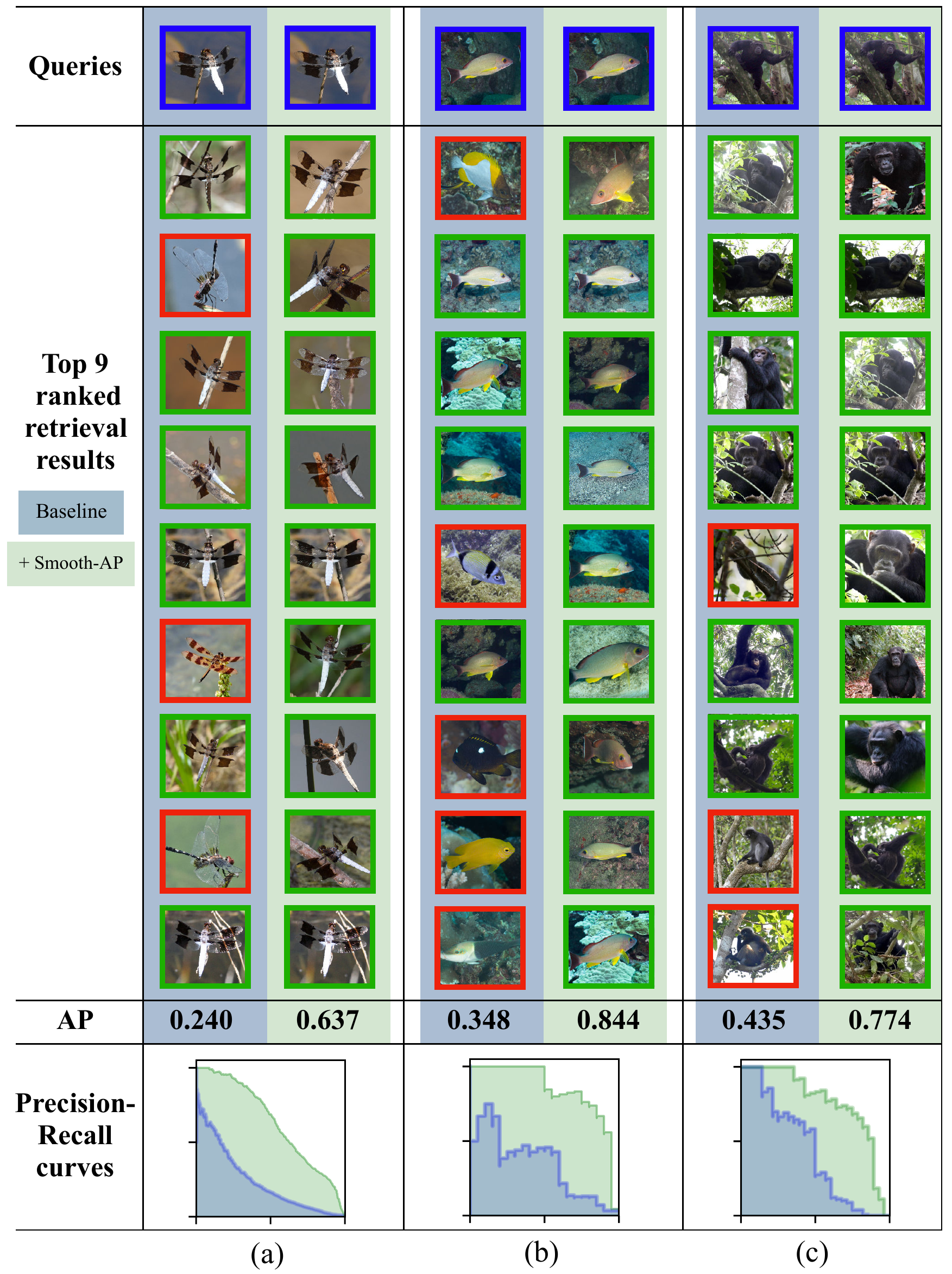}
\end{center}
\vspace{-10pt}
\beforecaptions
   \caption{\small{This Figure shows three separate query instances from the INaturalist dataset and the top ranked instances from the retrieval set when using the baseline model (ImageNet pre-trained weights), and after appending the Smooth-AP loss. As can be seen by the false positive retrieved instances for the baseline model, this is a highly challenging, fine-grained dataset; yet in all cases shown, appending the Smooth-AP loss leads to large gains in Average Precision. $|P|=$ 985, 20, 28 for (a),(b),(c) respectively.}}
 \aftercaptions
 \vspace{-1mm}
 \label{Inat}
\end{figure}

In this Section, we provide additional qualitative results for four of the datasets used in our experiments (VGGFace2 Test set, Stanford Online Products, and INaturalist). For each dataset, we display the top retrieved instances for various queries from the baseline model, both with and without appending the Smooth-AP loss.  
In all cases, the query example is shown in blue. The retrieved instances belonging to the same class as the query (\ie\ positive set) are shown in green, while the ones belonging to a different class from the query (\ie\ negative set) are shown in red. For all retrieval examples we have shown the corresponding precision-recall curves below, in which the baseline model is represented in blue, and the Smooth-AP model for the same query instance is represented overlaid in green. For Figures~\ref{face_1}, \ref{face_2} the retrieval set is ranked from left to right starting in the top row next to the query. For Figures~\ref{face_3}, \ref{face_4}, \ref{SOP}, \ref{Inat} the retrieval set is ranked from top to bottom. In each case, the Average Precision (AP) computed over the whole retrieval set is provided either below or alongside the ranked instances.

%% file: arxiv_sec/pseudo_code.tex
\newcommand\algcomment[1]{\def\@algcomment{\footnotesize#1}}
\begin{algorithm}[h]
	\caption{Pseudocode for Smooth-AP in Pytorch-style.}\label{fig:algo}
	\begin{lstlisting}[language=python]
# scores: predicted relevance scores (1 x m)
# gt: groundtruth relevance scores (1 x m)
 
 def t_sigmoid(tensor, tau=1.0):
     # tau is the temperature.
     exponent = -tensor / tau
     y = 1.0 / (1.0 + exp(exponent))
     return y
 
 def smooth_ap(scores, gt):
     # repeat the number row-wise.
     s1 = scores.repeat(m, 1) # s1: m x m
     # repeat the number column-wise.
     s2 = s1.transpose  # s2: m x m
     # compute difference matrix
     D = s1 - s2
     # approximating heaviside
     D_ = t_sigmoid(D, tau=0.01)
     # ranking of each instance
     R = 1 + sum(D_ * (1-eye(m)), 1)
     # compute positive ranking
     R_pos = gt.T * R
     # compute AP
     AP = (1 / sum(gt)) * sum(R_pos / R)
     return 1-AP
	\end{lstlisting}
\end{algorithm}

%% file: arxiv_sec/C.tex
\begin{figure}
\centering
\begin{subfigure}{.65\textwidth}
  \centering
  \includegraphics[width=.9\linewidth]{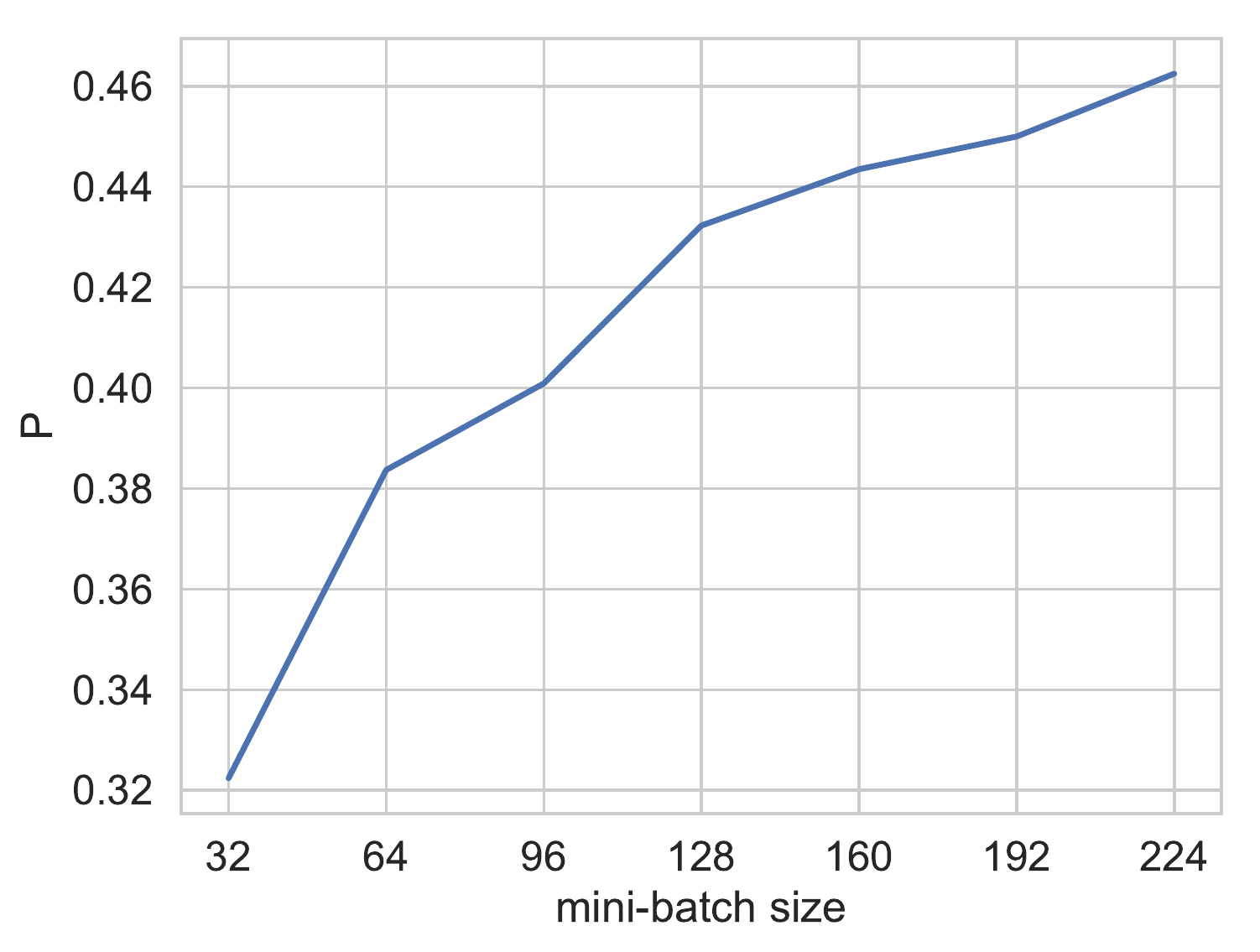}
  \caption{}
  \label{fig:sub1}
\end{subfigure}%
\begin{subfigure}{.35\textwidth}
  \centering
  \includegraphics[width=.9\linewidth]{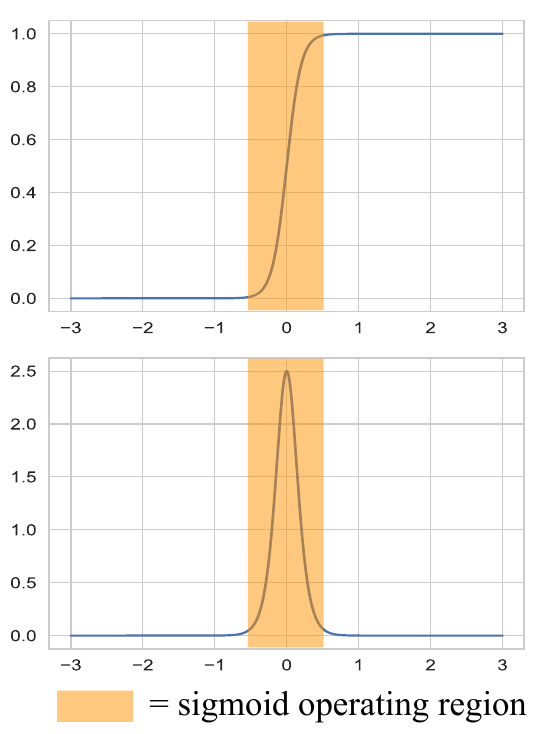}
  \caption{}
  \label{fig:sub2}
\end{subfigure}
\caption{\small{ (a): $P$, the proportion of elements of the difference matrix that fall into the operating region of the sigmoid (shown in (b)), and hence receive non negligible gradients, for several different mini-batch sizes. This explains why Smooth-AP benefits from large mini-batch sizes. }}
\label{fig:mini_batch_P}
\end{figure}

In this Section, we provide a further quantitative validation of the claims made in Section 6.4 (in the main manuscript) about the effects of mini-batch size on the Smooth-AP loss. We conjecture that a large mini-batch size increases the likelihood of relevance scores in the mini-batch being close to each other, and hence elements of the difference matrix (Equation 4 in main manuscript) falling into the narrow operating region of the sigmoid (we define the the operating region of the sigmoid as the narrow region with non-negligible gradients, see Figure~\ref{fig:sub2}), meaning that non-negligible gradients are fed backwards from Smooth-AP. This conjecture can be verified by increasing the mini-batch size during training and logging the proportion of elements of the difference matrix that fall into the operating region of the sigmoid. For each mini-batch during training, a difference matrix $D$ is constructed of size $(m * m)$ where $m$ is the mini-batch size. The proportion of elements of $D$ that fall into the operating region of the sigmoid used in Smooth-AP, which we denote as $P$,  can be computed using Equation~\ref{non_neg_grad} (we use a value of 0.005 to represent a non-negligible gradient). While keeping all parameters equal except mini-batch size, the average $P$ is computed across all mini-batches in one epoch of training on the Online Products dataset for several different mini-batch sizes, with the results plotted in Figure~\ref{fig:sub1}. As expected, $P$ increases with mini-batch size due to the fact that more instances in a mini-batch means that more instances are close enough together in terms of similarity score to lie within the operating zone of the sigmoid. This in turn leads to more non-negligible gradients being fed backwards to the network weights, and hence a higher evaluation performance, as was shown in the ablation Table 5 in the main manuscript.

\begin{equation}
\label{non_neg_grad}
P = \frac{\sum_{i=0}^{m-1} \sum_{j=0}^{m-1} (|\frac{d\mathcal{G}(D_{ij})}{dD_{ij}}{|} > 0.005)}{m^2} 
\end{equation}

%% file: arxiv_sec/D.tex
The two AP-optimising methods that we compare to for the INaturalist experiments (Table 3) have several hyper-parameters associated with them. 
For FastAP~\cite{Cakir19}, there is the number of histogram bins, $L$,
and for Blackbox AP~\cite{Rolnek20optimizing}, there is the value of $\lambda$ and the margin. 
For both methods we choose the hyper-parameters that are recommended in the respective publications for the largest dataset that was experimented on, which would be closest to INaturalist in terms of number of training images. For FastAP the number of histogram bins $L$ is set to 20, and for Blackbox AP, 
$\lambda$ is set to 4 and the margin is set to 0.02. We note that the evaluated Recall@K scores might be increased by varying these parameters. For all experiments on the INaturalist dataset, we multiply the learning rate on the last linear layer by a factor of 2.

%% file: arxiv_sec/A.tex


Table~\ref{tab:complexity} shows the time complexities of Smooth-AP, and also the other AP-optimising loss functions that we compare to. We also measure the times of the forward and backward passes for a single training iteration when each of the different loss functions are appended onto a ResNet50~\cite{he2016deep} backbone architecture. More specifically, we measure the forwards and backwards pass time for the backbone network, \textit{backbone time}, and the appended loss function, \textit{loss time}. These values for timings are averaged over all iterations in one training epoch for the Online Products dataset. The relevant notation: 
$\mathcal{M}$ is the number of instances in the retrieval set, which during mini-batch training is equal to the size of the mini-batch (with $p$ + $n$ = $\mathcal{M}$ and $p$, $n$ the number of positive and negative instances, respectively). 
$L$ refers to the number of bins used in the Histogram Binning technique~\cite{Cakir19,Ustinova16}. Even though the complexity of the proposed Smooth-AP loss is higher, Table~\ref{tab:complexity} shows that this leads to a very small overhead in computational cost on top of the ResNet50 backbone architecture ($ <3 ms $ for every iteration compared to previous methods where the backbone takes 705 ms), and hence in practice has a minor impact on the \textit{usability} of the proposed loss function. In all experiments here, all training parameters are equal ($|P|=4$, and mini-batch size $\mathcal{M}$ of 112).

\begin{table}[ht]
\begin{center}
\begin{tabular}{c|c|c|c}
\hline
{$Method$} & {$complexity$}& \textit{backbone time (ms)} & \textit{loss time (ms)} \\
\hline
Blackbox AP \cite{Rolnek20optimizing} &  $\mathcal{O}(n\log(n))$ & 705.0 & 3.7 \\
FastAP \cite{Cakir19} & $\mathcal{O}(\mathcal{M}L)$ & 705.0 & 4.2 \\
Smooth-AP & $\mathcal{O}(\mathcal{M}^2)$ & 705.0 & 6.6 \\
\hline
\end{tabular}
\end{center}
\caption{\small{The time complexities of different AP-optimising methods that we compare to, as well as the time taken for the forwards and backwards pass through the backbone for one iteration, \textit{backbone time}, and the time taken for the computation of the loss, \textit{loss time}. The slightly increased time complexity of Smooth-AP leads to a negligible increase in training time.}}
\label{tab:complexity}
\aftercaptions
\end{table}